# Can virtual reality predict body part discomfort and performance of people in realistic world for assembling tasks?

Bo HU[a], Liang MA[b], Wei ZHANG[a], Gaverial. SALVENDY[a], Damien CHABLAT[b], Fouad BENNIS[b]

a: Department of Industrial Engineering, Tsinghua University, Beijing,100084, P.R. China

b: Institut de Recherche en Communications et en Cybernétique de Nantes, CNRS UMR 6597

Ecole Centrale de Nantes, IRCCyN - 1, rue de la Noë - BP 92 101 - 44321

**Abstract**

This paper presents our work on relationship of evaluation results between virtual environment (VE) and realistic environment (RE) for assembling tasks. Evaluation results consist of subjective results (BPD and RPE) and objective results (posture and physical performance). Same tasks were performed with same experimental configurations and evaluation results were measured in RE and VE respectively. Then these evaluation results were compared. Slight difference of posture between VE and RE was found but not great difference of effect on people according to conventional ergonomics posture assessment method. Correlation of BPD and performance results between VE and RE are found by linear regression method. Moreover, results of BPD, physical performance, and RPE in VE are higher than that in RE with significant difference. Furthermore, these results indicates that subjects feel more discomfort and fatigue in VE than RE because of additional effort required in VE.

**Relevance to industry**

With digital mock-up and VR simulation, work design is evaluated to find potential ergonomics problems at early design stage of works in industries. It reduces cost as well as time consuming. The difference and correlation of evaluation results between VE and RE provide a reference for this method in work design.

# 1. Introduction

Musculoskeletal disorder (MSD) is one of serious occupational healthy problems to manual handling workers in industrialized countries, and it affects a significant proportion of workforce. In 2001, National Research Council and Institute of Medicine reported that MSD represented 40% of compensated injuries and cost between 45 and 54 billion dollars per year in United States(NationalResearchCouncil/InstituteofMedicine, 2001). In European Union, there was 40 millions workers suffering from MSD and the financial loss caused by MSD was about between 0.2% and 5% GDP by some estimation(Buckle and Devereux, 1999). Hence many researches focus on analyzing potential MSD exposures and how to prevent MSD in work design.

It is believed that MSD is closely related to postures, physical overexertion, duration and frequency of physical effort, discomfort, and physical fatigue (Pheasant, 1999). In order to prevent MSD risks, many evaluation methods have been developed to investigate ergonomics problems in design. These methods can be mainly classified into subjective (e.g., RPE, BPD) and objective evaluation methods (e.g., RULA, OWAS)(Li and Buckle, 1999). Borg's scale, also called Rated Perceiving Exertion (RPE) method, has been applied for evaluating effort of subjects in variety of researches and it has been validated in consistence to several psychological variables (e.g., heart rate) (Garcin, Vautier et al., 1998; Kim, Martin et al., 2004). Body part discomfort (BPD) method was developed to evaluate discomfort intensity of subjects. (Corlett and Bishop, 1976; Lowe, 2004; Yuan and Kuo, 2006). Some posture-based observation methods have been developed to assess physical exposures objectively. OWAS was designed to facilitate evaluation process of the overall human body (Scott and Lambe, 1996). Posture targeting method (Corlett, Madeley et al., 1979) and REBA (Hignett and McAtamney, 2000) were designed to evaluate entire body postures, while RULA was specially designed to evaluate upper body postures (McAtamney and Corlett, 1993; Bao, Howard et al., 2007). However, in these conventional methods, evaluation has to be carried out in field and it requires much effort and expensive physical mock-up.

Digital human modeling and virtual human simulation (e.g., 3DSSPP™, EAI

Jack®, RAMSIS) have been created to facilitate ergonomic evaluations. Using these tools, visual scope and reach envelope of users representing specific populations can be analyzed (e.g. EAI Jack®)(Chaffin, Nelson et al., 2001). Some DHM tools can calculate out biomechanics attributes of manual handling operations (e.g. Anybody® Modeling System, 3DSSPP™) and give predictions of fatigue and disorder. These analysis results can be used to find and fix ergonomics problems of proposed designs and improve the work design.

Virtual human simulation provides a quick, virtual representation of human being in simulated working environment. Physical mock-up is not any more necessary in virtual human simulation, and different aspects can be assessed with rapid computational efficiency. The main issue of using virtual human simulation is that the movement or the motion is obtained via inverse kinematics, and the virtual human has a robot-like behavior, but not natural or mimic enough (Chaffin and Erig, 1991).

Using virtual reality (VR) technology is able to provide an immersive working environment. Several peripheral devices have been invented to provide different interaction ways between user and VR systems, such as: motion tracking systems, haptic interfaces, etc. VR techniques, in combination with DHM tools, have been used more and more to enable the participation of human being (Buck, 1998). VR has also been used in ergonomic applications (Whitman, Jorgensen et al., 2004; Jayaram, Jayaram et al., 2006; Wang, Liao et al., 2007).

The aim of integrating ergonomic evaluation methods into VR is to facilitate work design process, enhance design efficiency, and lower the design cost. Hypothetically, if a virtual environment could provide 100% fidelity, the workload in virtual environment (VE) might be the same as in realistic environment (RE). Our main concern is whether the evaluation result in VR with different presence level is consistent to the evaluation in RE. Therefore, we proposed an experimental approach to check the relation between the evaluation results in VR and VE for same physical operations.

In this paper, we presented our preliminary experiment results in a VR system only with visual feedback. The purpose of our research is to analyze evaluation results in VE and RE for the same tasks. Subjective evaluation methods (BPD, RPE) are used to evaluate discomfort and perceived exertion effort in both environments, and objective methods (posture, fatigue) are used to evaluate physical aspects of the task. The relationship of evaluation results in RE and VE is analyzed with regression method.

## 2. Method

*2.1. Subjects*

30 male subjects participated in this experiment after giving their informed consent. They were all recruited from a manufacturing enterprise. They were all free from musculoskeletal disorders. Their mean age was 41.8 *y* (s.d.=11.5 *y*), mean height was 172.0 *cm* (s.d.=5.5 *cm*), and the mean mass was 69.5 *kg* (s.d.=12.2 *kg*). Twenty-two of them are all professional hand-tool operators and their working experience varies from 3 to 20 years. The other subjects use hand-tools occasionally.

*2.2. Task description*

The task in this experiment was designed by simplified from typical assembling tasks (hand drilling operations). The task consisted of several elementary operations in assembling tasks: holding and lifting a hand-tool, reaching and hitting targets, and keeping alignment between the tool and the target for assembly operations. Each subject was asked to perform the simulated drilling operations while sitting or standing at a fixed working position and facing at a tool work station and a work platform.

A 1.5 *kg* weighted hand-tool was used in this experiment for simulating the external physical load. The hand-tool was made of a plastic cover with weighted materials to replace a real powered pistol drill. It was placed on a fixed work station, which had a height of 70 *cm* and was placed in front of the subject with a horizontal distance 80 *cm*.

The working position of a subject was fixed in the workplace. Operation targets were placed on two fixed platforms, for sitting posture and standing posture, respectively. Both platforms were placed 80 *cm* ahead of the subject. The heights of the platforms were 80 *cm* and 140 *cm,* for sitting posture and standing posture, respectively.

All the operation targets were located in target models. The size of target model for standing tasks was 550 *mm* (Height) by 400 *mm* (Width), and the size of target model for sitting tasks was 400 *mm* (Height) by 600 *mm* (Width). There were 9 numerated target points with different positions. Physical target models were used in RE, while the same digital models were used in VE (Figure 1)

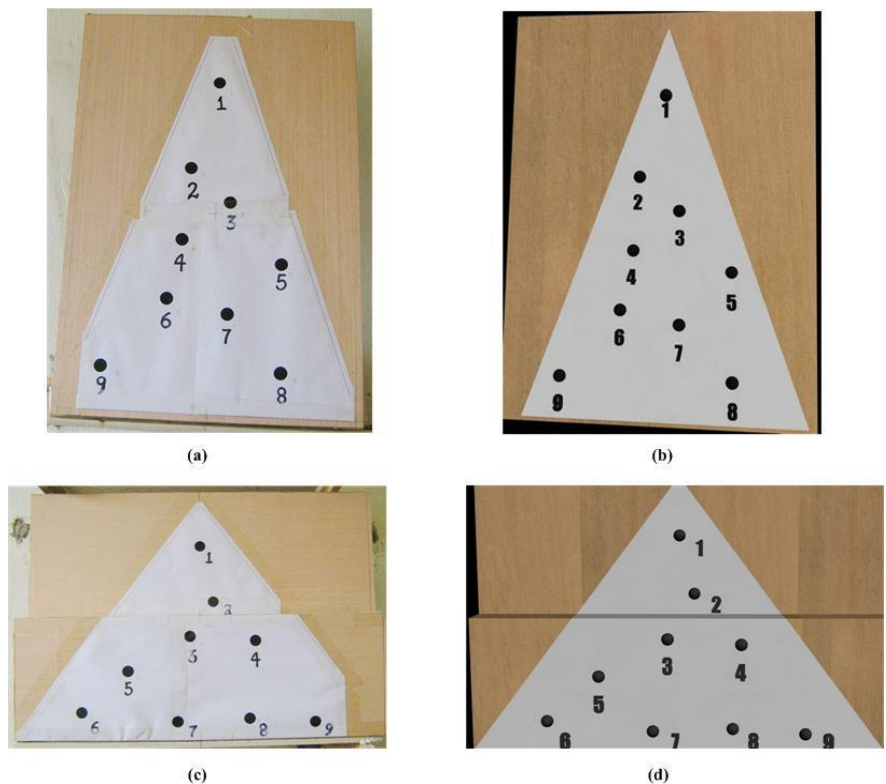

(a): physical model for standing task.　(b): digital model for standing task
(c): physical model for sitting task.　(d): digital model for sitting task
Figure 1　The target models in VE and RW

The following steps were necessary to complete simulated assembling tasks.

1. The subject reached and held the hand-tool at a fixed position and lifted it.

2. The subject placed the hand-tool in alignment with a target point of the model and kept his posture for 4 seconds.

3. The subject had to finish all the 9 target points in ascending order as a loop.

4. One task might include different number of loops.

*2.3. Apparatus*

The VR simulation system used in this experiment was developed in the Virtual Reality and Human Interface Technology (VRHIT) laboratory of Tsinghua University. This system provided immersive virtual scenarios for subjects based on OpenGL®. Virtual objects can be created with 3D modeling tools (e.g. AutoDesk® 3ds MAX, PTC® Pro/E) and imported into the system. Motion tracking devices, digital head mounted display (HMD), data gloves, and multimodal feedback devices can be linked to the system as peripherals.

Meanwhile, a manikin (digital human model) can be provided in this VR simulation system, and it can be driven by captured data from motion tracking devices. The manikin interacts with other virtual objects, and interactive virtual scenes are displayed to subjects. The manikin is used to provide visual representation of the subject in VE for better understanding the interactions between subject and virtual scenarios.

The VR motion simulation system can be used for simulating assembling, handling, and maintenance operations in industries. It has been used in projects collaborated with industrial enterprises (Hu, Wang et al., 2008).

In this experiment, we used two sets of magnetic motion tracking devices (totally six sensors) made by Pohemus Corp to track subject's motion. These sensors were mounted on the head and key joints of a subject for tracking working posture. The acquisition rate is 30 *Hz* per sensor and the static accuracy of each sensor is 1 *mm*.

Taking a right-handed worker as an example, four electromagnetic sensors were mounted on the head, the right shoulder, the right elbow, and the right wrist of a subject (Figure 2). The captured coordinates by sensors represented the positions of key joints and depicted the posture of the subject. In this experiment, the sensor mounted on head was used to determine position and orientation of viewpoint in VE. Sensors mounted on shoulder, elbow, and wrist, were used to calculate the posture of right arm. The other two sensors were mounted on the two ends of the hand-tool to obtain its position and orientation while subjects were performing tasks.

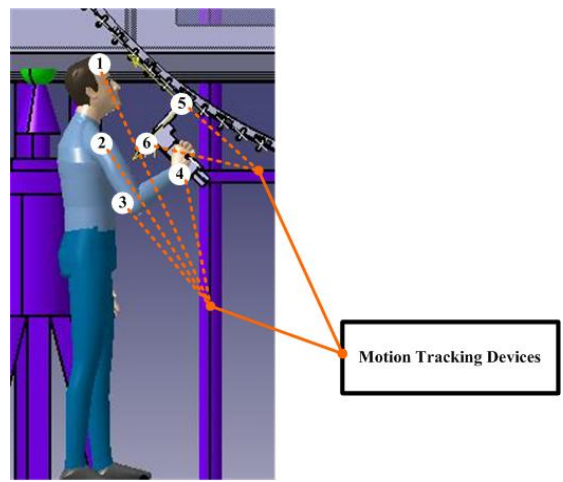

Figure 2 Magnetic sensors mounted on a subject

A digital HMD, made by 5DT Corp., was used for displaying immersive virtual scenarios to subjects in the experiment. Its field of view (FOV) was 39° (Horizon) by 31° (Vertical) and its resolution was 800 by 600 (pixels).

A force measurement device was used in the experiment for measuring the force capacity of a subject's arm. Its maximum measuring range is 60 *kg*. Its resolution was 0.01 *kg*.

The force measurement device and motion tracking devices were used in RE as well.

*2.4. Design of experiment*

The objective of this experiment is to compare evaluation results between RE

and VE, then find out their relationship and difference. Considering possible effects on experiments, three with-subjects factors were involved in this experiment. Their levels and descriptions are demonstrated in Table 1.

All tasks were done both in RE and VE. Subjects were divided into two groups for different sequences of sessions. Group 1 performed the RE session at first, then the VE session. With an inverse sequence, Group 2 performed the VE session at first, then the RE session. Tasks of RE sessions were the same as those in VE sessions, and the relative position of a target model to a subject in the RE sessions were the same as those in VE sessions.

Table 1 Factors and their levels in the experiment

| Factors | Levels | Description |
| --- | --- | --- |
| Working posture tasks | ST | Sitting task |
|  | SD | Standing task |
| Duration of tasks | L | 4 loops in a task |
|  | S | 2 loops in a task |
| Size of target point | LG | Diameter of a target point is 6 *mm* |
|  | SL | Diameter of a target point is 18 *mm* |

Working posture tasks, duration of tasks, and size of target points were involved as within-subject factors in the experiment. Each factor has two levels. Working posture tasks consist of sitting posture and standing posture. Two types of target points with different sizes were designed initially to represent the difficulty of an assembly task in this experiment: ø6 and ø18. The duration of tasks has two levels: one level is two loops in a task and the other level is four loops in a task.

These three factors make up of eight different treatments for a subject. A performing sequence of these treatments was assigned randomly to each subject. Each subject performs tasks with the same sequence of treatments in RE session and VE session.

*2.5. Experimental protocol*

*2.5.1. Objective evaluation results*

Two objective indexes, working posture and physical performance (fatigue), were measured for assessing operation tasks.

In our experiment, in most cases, right arm was mainly engaged in the operation, and the angle between the forearm and the upper arm indicated the working posture directly. This variable could be used to assess the posture.

The physical performance of each subject, indicated by the decrease of the maximum force capacity, was measured to representing fatigue since the decrease of the force capacity is the most direct measurement of fatigue (Vøllestad, 1997). The maximum voluntary contraction of the right arm was measured by the force measurement device and denoted as $F_{before}$ before starting a task. After finishing a task, the maximum voluntary contraction was measured again and it was denoted as $F_{after}$. The normalized decrease of force capacity, calculated by $(F_{before} - F_{after})/F_{before}$, was used as physical performance in this experiment.

*2.5.2. Subjective evaluation results*

The subjective evaluation was carried out using two methods: BPD and RPE. A self-reported questionnaire, consisting of BPD scale and RPE scale, was used for collecting subjective sensation of subjects in RE and VE. In the questionnaire, Borg's scale with 6-20 point was used for RPE questions. The 11-point scale and the original body part diagram of Corlett was used in BPD scale (Corlett and Bishop, 1976). The original body part diagram of BPD divided entire body into twelve regions: neck, shoulders, upper arms, forearms, upper back, middle back, lower back, buttocks, left thigh, right thigh, left shank, right shank.

Subjects were asked to report their evaluation of BPD and RPE when they finished a task as subjective evaluation result.

*2.5.3. Presence of VE*

A presence questionnaire was used for assessing presence of a subject in VE

session. The presence questionnaire of this experiment was created referring to Witmer and Singer's presence questionnaire (Witmer and Singer, 1998). Seven-point scale was used in the presence questionnaire. Higher total score of answers in scale of the questionnaire indicates higher presence.

## 2.6. Data analysis

### 2.6.1. Posture

The posture of each subject was denoted as an 8 by 9 matrix, $A$. Each row represents one of the eight experimental configurations and each column represents one of the nine points on target models. The eight experiment configurations were enumerated in Table 2. For each element of $A$, denoted as $a_{ij}$, represents the posture under a given experiment configurations while operating a given target point.

Table 2 Experimental configurations represented by the subscript i

| $i$ | **Experimental Configuration** |
|---|---|
| 1 | ST and SL and L |
| 2 | ST and SL and S |
| 3 | ST and LG and L |
| 4 | ST and LG and S |
| 5 | SD and SL and L |
| 6 | SD and SL and S |
| 7 | SD and LG and L |
| 8 | SD and LG and S |

The relative posture difference matrix between VE and RE, denoted as $A'$, can be calculated by $a'_{ij} = \left(a_{ij}^{RE} - a_{ij}^{VE}\right)/a_{ij}^{RE}$, which is also an 8 by 9 matrix. Furthermore, the posture difference for each subject between RE and VE was denoted as a vector, $\Delta$, whose element was calculated by $\delta_i = \frac{1}{9}\sqrt{\sum_{i=1}^{9}\left(a'_{ij}\right)^2}$.

### 2.6.2. Physical performance

The physical performance, calculated by $\left(F_{before} - F_{after}\right)/F_{before}$, was used as fatigue index of subjects. Performance of thirty subjects was denoted as an 8 by 30 matrix $P$. Each row represents one of the eight experimental configurations and each column represents one of the thirty subjects. For each element of $P$, denoted as $p_{ij}$, experimental configurations represented by the subscript $i$ was demonstrated in Table 2. Then some matrices of physical performance in a working environment can be calculated out by categorized with different factors (Table 3).

Table 3  The physical performance categorized with different factors

| Term | Calculation | Description |
|---|---|---|
| $P'$ | $p'_j = \dfrac{1}{8}\sum_{i=1}^{8} p_{ij}$ | The overall mean of a working environment |
| $P^{SD}$ | $p^{SD}_j = \dfrac{1}{4}\sum_{i=1}^{4} p_{ij}$ | The mean of standing posture tasks |
| $P^{ST}$ | $p^{ST}_j = \dfrac{1}{4}\sum_{i=5}^{8} p_{ij}$ | The mean of sitting posture tasks |
| $P^{L}$ | $p^{L}_j = \dfrac{1}{4}(p_{1j} + p_{3j} + p_{5j} + p_{7j})$ | The mean of long duration tasks |
| $P^{S}$ | $p^{S}_j = \dfrac{1}{4}(p_{2j} + p_{4j} + p_{6j} + p_{8j})$ | The mean of short duration tasks |
| $P^{LG}$ | $p^{LG}_j = \dfrac{1}{4}(p_{3j} + p_{4j} + p_{7j} + p_{8j})$ | The mean of tasks with large size points of target model |
| $P^{SL}$ | $p^{SL}_j = \dfrac{1}{4}(p_{1j} + p_{2j} + p_{5j} + p_{6j})$ | The mean of tasks with small size points of target model |

Difference and relationship of physical performance were analyzed with these results of RE and corresponding results of VE with paired t-test and linear regression method respectively.

*2.6.3. Body part discomfort*

BPD method measures discomfort intensities of 12 body parts in the entire body. For the $k^{th}$ part of body, the BPD results were denoted as an 8 by 30 matrix $B_k$. The element of the $B_k$, $b_{ij}$, represents the BPD result of the the $j^{th}$ subject under the $i^{th}$ experimental configuration. Then some matrices of BPD result can be obtained by categorized with different factors (Table 4).

Table 4 The BPD categorized with different factors

| Term | Calculation | Description |
|---|---|---|
| $B'_k$ | $b'_j = \frac{1}{8}\sum_{i=1}^{8} b_{ij}$ | The overall mean of a working environment |
| $B_k^{SD}$ | $b_j^{SD} = \frac{1}{4}\sum_{i=1}^{4} b_{ij}$ | The mean of standing posture tasks |
| $B_k^{ST}$ | $b_j^{ST} = \frac{1}{4}\sum_{i=5}^{8} b_{ij}$ | The mean of sitting posture tasks |
| $B_k^{L}$ | $b_j^{L} = \frac{1}{4}(b_{1j} + b_{3j} + b_{5j} + b_{7j})$ | The mean of long duration tasks |
| $B_k^{S}$ | $b_j^{S} = \frac{1}{4}(b_{2j} + b_{4j} + b_{6j} + b_{8j})$ | The mean of short duration tasks |
| $B_k^{LG}$ | $b_j^{LG} = \frac{1}{4}(b_{3j} + b_{4j} + b_{7j} + b_{8j})$ | The mean of tasks with large size points of target model |
| $B_k^{SL}$ | $b_j^{SL} = \frac{1}{4}(b_{1j} + b_{2j} + b_{5j} + b_{6j})$ | The mean of tasks with small size points of target model |

For BPD data analysis, two rules were used for eliminating outlier data before data analysis. The first rule was that the BPD scores of RE and VE less than 1-point were eliminated. 1-point is assigned as very little discomfort sensation and 0.5-point is assigned as just feel discomfort sensation in scale of BPD. Using 1-point as a threshold for filtering outlier data is helpful to decrease unstable data of BPD. The second rule was that BPD data were eliminated when BPD scores for a subject were all zero in RE and VE. 0-point is assigned as no discomfort sensation in scale of BPD. Hence these kinds of data are eliminated as outlier data is helpful to decrease interference of void data. After eliminating data, results of 27 subjects were used for analysis.

Difference and relationship of BPD were analyzed with these results of RE and corresponding results of VE with paired t-test and linear regression method respectively.

*2.6.4. Rated perceived exertion(RPE)*

RPE method was used to measure effort of subjects for performing tasks. RPE result of thirty subjects was denoted as an 8 by 30 matrix $R$. The element of $R$, denoted as $r_{ij}$, represented the RPE result of the $j^{th}$ subject under the $i^{th}$ experimental configuration. Then some matrices can be calculated out by categorized with different factors in a working environment (Table 5).

Table 5 The RPE categorized with different factors

| Term | Calculation | Description |
|---|---|---|
| $R'$ | $r'_j = \frac{1}{8}\sum_{i=1}^{8} b_{ij}$ | The overall mean of a working environment |
| $R^{SD}$ | $r^{SD}_j = \frac{1}{4}\sum_{i=1}^{4} r_{ij}$ | The mean of standing posture tasks |
| $R^{ST}$ | $r^{ST}_j = \frac{1}{4}\sum_{i=5}^{8} r_{ij}$ | The mean of sitting posture tasks |
| $R^{L}$ | $r^{L}_j = \frac{1}{4}(r_{1j} + r_{3j} + r_{5j} + r_{7j})$ | The mean of long duration tasks |
| $R^{S}$ | $r^{S}_j = \frac{1}{4}(r_{2j} + r_{4j} + r_{6j} + r_{8j})$ | The mean of short duration tasks |
| $R^{LG}$ | $r^{LG}_j = \frac{1}{4}(r_{3j} + r_{4j} + r_{7j} + r_{8j})$ | The mean of tasks with large size points of target model |
| $R^{SL}$ | $r^{SL}_j = \frac{1}{4}(r_{1j} + r_{2j} + r_{5j} + r_{6j})$ | The mean of tasks with small size points of target model |

Difference of RPE was analyzed with these results of RE and corresponding results of VE with paired t-test.

Both objective and subjective evaluation results were analyzed with SPSS® and analysis results were plotted with Sigmaplot®.

## 3. Results

*3.1. Posture*

Difference of a subject's posture between RE and VE was denoted by an eight by one vector $\Delta$, which represents posture difference under different experimental configuration. Hence the overall posture difference of 30 subjects was denoted as a 240 by 1 matrix $\Theta = (\Delta_1^T \cdots \Delta_{30}^T)^T$, representing all observed value of thirty subjects. The histogram of the observed differences was plotted in Figure 3, and an interval 5% is used to divide all the observed differences.

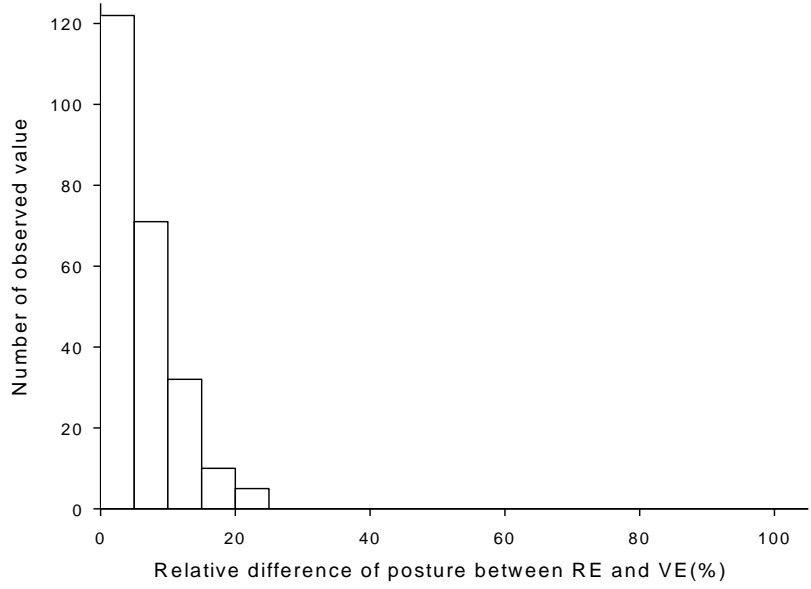

Figure 3 Histogram of relative difference of posture between RE and VE

It was observed that 80.4% of 240 observed differences were in the interval [0%, 5%], and 93.8% in the interval [0%, 10%].

*3.2. Physical performance (Fatigue)*

The physical performance result of 30 subjects measured in RE was denoted as $P'_{RE}$, and that measured in VE was denoted as $P'_{VE}$. The mean of $P'_{RE}$, which was calculated by $\overline{p'_{RE}} = \frac{1}{30}\sum_{j=1}^{30} p'_{RE_j}$, was 0.126 (s.d.=0.03795). The mean of $P'_{VE}$, which was calculated by $\overline{p'_{VE}} = \frac{1}{30}\sum_{j=1}^{30} p'_{VE_j}$, was 0.147 (s.d.=0.0454). Difference between $P'_{RE}$ and $P'_{VE}$ was tested with paired t-test, and significant difference was found ($t_{RE-VE} = -3.838$, $p < 0.05$).

Furthermore $P'_{RE}$ and $P'_{VE}$ were regressed by a linear model, and result was ($R^2 = 0.568$, $F = 36.835$, $p < 0.05$):

$$\hat{p}'_{RE} = 0.033 + 0.629\,\hat{p}'_{VE}$$

The regression was shown in Figure 4.

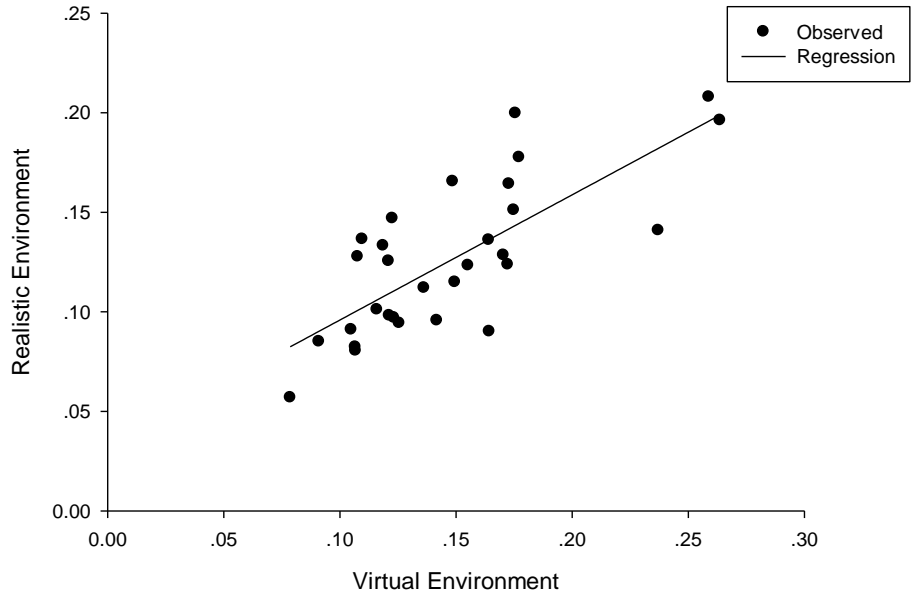

Figure 4 Regression result of the physical performance

Performance results were classified and analyzed under different experimental configurations. All results were classified into different categories for analysis based on three factors: working posture tasks, duration of tasks, and size of target points.

*3.2.1. Working Postures Tasks: Sitting vs. Standing*

In this experiment, working posture tasks consist of standing and sitting posture tasks. Performance result of standing posture tasks, measured in RE, was denoted as $P_{RE}^{SD}$, and that measured in VE was denoted as $P_{VE}^{SD}$. Performance result of sitting posture tasks, measured in RE, was denoted as was denoted as $P_{RE}^{ST}$, and that measured in VE was denoted as $P_{VE}^{ST}$. The descriptive information of $P_{RE}^{SD}$, $P_{VE}^{SD}$, $P_{RE}^{ST}$, and $P_{VE}^{ST}$ was demonstrated in Table 6.

Table 6 The descriptive information of $P_{RE}^{SD}$, $P_{VE}^{SD}$, $P_{RE}^{ST}$, and $P_{VE}^{ST}$

| Working Posture Task | Performance | Mean | Std Deviation |
|---|---|---|---|
| **Standing Tasks** | $P_{RE}^{SD}$ | 0.123 | 0.03782 |

|  | | Mean | Std Deviation |
|---|---|---|---|
| | $P_{VE}^{SD}$ | 0.151 | 0.05599 |
| **Sitting Tasks** | $P_{RE}^{ST}$ | 0.129 | 0.04375 |
| | $P_{VE}^{ST}$ | 0.144 | 0.04003 |

Difference between $P_{RE}^{SD}$ and $P_{VE}^{SD}$ was tested with paired t-test and the significant difference was found ($t_{RE-VE} = -3.680$, $p < 0.05$). Furthermore $P_{RE}^{SD}$ and $P_{VE}^{SD}$ were regressed by a linear model and its result was ($R^2 = 0.462$, $F = 24.011$, $p < 0.05$):

$$\hat{p}_{RE}^{SD} = 0.054 + 0.459\, \hat{p}_{VE}^{SD}$$

Difference between $P_{RE}^{ST}$ and $P_{VE}^{ST}$ was tested with paired t-test and significant difference was found ($t_{RE-VE} = -2.671$, $p < 0.05$). $P_{RE}^{ST}$ and $P_{RE}^{ST}$ were regressed by a linear model and its result was ($R^2 = 0.561$, $F = 35.798$, $p < 0.05$):

$$\hat{p}_{RE}^{ST} = 0.011 + 0.819\, \hat{p}_{VE}^{ST}$$

*3.2.2. Duration of Task: Long vs. Short*

In this experiment, duration of task consists of long and short duration. Performance result of long duration tasks, measured in RE, was denoted as $P_{RE}^{L}$, and that measured in VE was denoted as $P_{VE}^{L}$. Performance result of short duration tasks, measured in RE, was denoted as $P_{RE}^{S}$, and that measured in VE was denoted as $P_{VE}^{S}$. The descriptive information of $P_{RE}^{L}$, $P_{VE}^{L}$, $P_{RE}^{S}$, and $P_{VE}^{S}$ was demonstrated in Table 7.

Table 7 The descriptive information of $P_{RE}^{L}$, $P_{VE}^{L}$, $P_{RE}^{S}$, and $P_{VE}^{S}$

| Duration of Task | Performance | Mean | Std Deviation |
|---|---|---|---|
| **Long** | $P_{RE}^{L}$ | 0.147 | 0.04447 |
| | $P_{VE}^{L}$ | 0.174 | 0.05697 |
| **Short** | $P_{RE}^{S}$ | 0.105 | 0.03653 |
| | $P_{VE}^{S}$ | 0.121 | 0.03947 |

Difference between $P_{RE}^{L}$ and $P_{VE}^{L}$ was tested with paired t-test and the significant difference was found ($t_{RE-VE} = -3.376$, $p < 0.05$). Furthermore $P_{RE}^{L}$ and $P_{VE}^{L}$ were regressed by a linear model and its result was ($R^2 = 0.450$, $F = 22.901$, $p < 0.05$):

$$\hat{p}_{RE}^{L} = 0.056 + 0.524 \, \hat{p}_{VE}^{L}$$

Difference between $P_{RE}^{S}$ and $P_{VE}^{S}$ was tested with paired t-test and significant was found ($t_{RE-VE} = -2.665$, $p < 0.05$). $P_{RE}^{S}$ and $P_{VE}^{S}$ were regressed by a linear model and its result was ($R^2 = 0.404$, $F = 18.965$, $p < 0.05$):

$$\hat{p}_{RE}^{S} = 0.034 + 0.588 \, \hat{p}_{VE}^{S}$$

### 3.2.3. Size of the target points: Large vs. Small

In this experiment, size of target points consists of large and small target points. The performance of large target points tasks measured in RE was denoted as $P_{RE}^{LG}$, and that measured in VE was denoted as $P_{VE}^{LG}$. The performance of small target points tasks measured in RE was denoted as $P_{RE}^{SL}$, and that measured in VE was denoted as $P_{VE}^{SL}$. The descriptive information of $P_{RE}^{LG}$, $P_{VE}^{LG}$, $P_{RE}^{SL}$, and $P_{VE}^{SL}$ was demonstrated in Table 8.

Table 8 The descriptive information of $P_{RE}^{LG}$, $P_{VE}^{LG}$, $P_{RE}^{SL}$, and $P_{VE}^{SL}$

| Size of target points | Performance | Mean | Std Deviation |
|---|---|---|---|
| **Large** | $P_{RE}^{LG}$ | 0.125 | 0.04543 |
|  | $P_{VE}^{LG}$ | 0.149 | 0.05140 |
|  | $P_{RE}^{SL}$ | 0.127 | 0.03648 |
| **Small** |  |  |  |
|  | $P_{VE}^{SL}$ | 0.145 | 0.04233 |

Difference of $P_{RE}^{LG}$ and $P_{VE}^{LG}$ was tested with paired t-test and significant difference was found ($t_{RE-VE} = -3.528$, $p < 0.05$). Furthermore $P_{RE}^{LG}$ and $P_{VE}^{LG}$ were regressed by a linear model and its result was ($R^2 = 0.504$, $F = 28.421$, $p < 0.05$):

$$\hat{p}_{RE}^{LG} = 0.032 + 0.627\, \hat{p}_{VE}^{LG}$$

Difference of $P_{RE}^{SL}$ and $P_{VE}^{SL}$ was tested with paired t-test and significant difference was found ($t_{RE-VE} = -3.197$, $p < 0.05$). $P_{RE}^{SL}$ and $P_{VE}^{SL}$ were regressed by linear model and its result was ($R^2 = 0.465$, $F = 24.367$, $p < 0.05$):

$$\hat{p}_{RE}^{SL} = 0.041 + 0.588\, \hat{p}_{VE}^{SL}$$

*3.3. Body part discomfort*

Discomfort of forearms, the upper arms, the shoulders, and neck was reported by subjects in this experiment. However, only significant correlation of forearms between RE and VE was found.

For forearms, BPD result of thirty subjects was denoted as $B'$. These BPD results of $B'$ were filtered according to eliminating rules mentioned above, and remained 27 elements of $B'$ were used for analysis. The BPD result of forearms, measured in RE, was denoted as $B'_{RE}$, and that measured in VE was denoted as $B'_{VE}$.

The mean of $B'_{RE}$ was 2.242 (s.d.=1.285), and the mean of $B'_{VE}$ was 3.290 (s.d.=1.070). Paired T-test was used for testing difference of $B'_{RE}$ and $B'_{VE}$, and their difference was significant ($t_{RE-VE} = -5.932$, $p < 0.05$). $B'_{RE}$ and $B'_{VE}$ were regressed by a linear model, and the regression result was ($R^2 = 0.506$, $F = 25.561$, $p < 0.05$):

$$\hat{b}'_{RE} = -0.567 + 0.854\, \hat{b}'_{VE}$$

The result was shown in Figure 5.

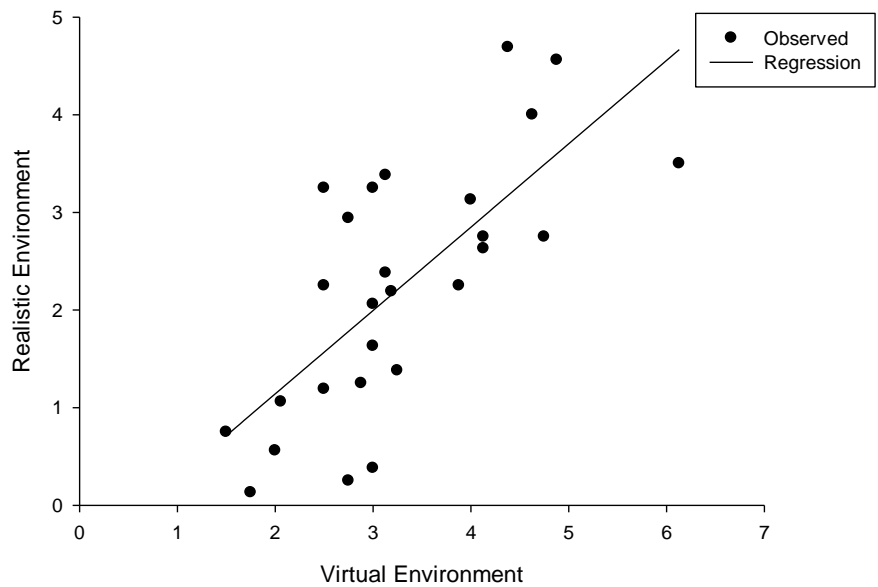

Figure 5 The regression result of $B'_{RE}$ and $B'_{VE}$

BPD results of forearms were classified and analyzed by different experimental configurations. All results were classified into different categories for analysis based on three factors: working posture tasks, duration of tasks, and size of target points.

*3.3.1. Working posture: Standing vs. Sitting*

In factors of this experiment, working posture tasks consist of sitting and standing posture tasks. For forearms, The BPD result of standing posture, measured in RE, was denoted as $B^{SD}_{RE}$, and that measured in VE was denoted as $B^{SD}_{VE}$. The BPD result of sitting posture tasks, measured in RE, was denoted as $B^{ST}_{RE}$, and that measured in VE was denoted as $B_{VE}^{ST}$.

Observed value of $B_{RE}^{SD}$, $B_{VE}^{SD}$, $B_{RE}^{ST}$, and $B_{VE}^{ST}$ was filtered by two rules and 27 elements of each matrix was remained for analysis. The descriptive information of $B_{RE}^{SD}$, $B_{VE}^{SD}$, $B_{RE}^{ST}$, and $B_{VE}^{ST}$ was demonstrated in Table 9.

Table 9 The descriptive information of $B_{RE}^{SD}$, $B_{VE}^{SD}$, $B_{RE}^{ST}$, and $B_{VE}^{ST}$

| Working Posture Tasks | BPD | Mean | Std Deviation |
|---|---|---|---|
| Standing tasks | $B_{RE}^{SD}$ | 2.139 | 1.168 |
| | $B_{VE}^{SD}$ | 3.301 | 1.339 |
| Sitting tasks | $B_{RE}^{ST}$ | 2.353 | 1.621 |
| | $B_{VE}^{ST}$ | 3.273 | 1.045 |

Paired T-test was performed for finding difference of $B_{RE}^{SD}$ and $B_{VE}^{SD}$. Significant difference was found between them ($t_{RE-VE} = -5.266$, $p < 0.05$). Furthermore linear regression method was used for $B_{RE}^{SD}$ and $B_{VE}^{SD}$ with the linear model. The regression result was ($R^2 = 0.347$, $F = 13.280$, $p < 0.05$):

$$\hat{b}_{RE}^{SD} = 0.442 + 0.514\ \hat{b}_{VE}^{SD}$$

Paired T-test was performed for testing difference of $B_{RE}^{ST}$ and $B_{VE}^{ST}$. Significant difference was found for them ($t_{RE-VE} = -4.145$, $p < 0.05$). Furthermore regression method was used for $B_{RE}^{ST}$ and $B_{VE}^{ST}$ with linear model. The regression result was ($R^2 = 0.485$, $F = 23.529$, $p < 0.05$):

$$\hat{b}_{RE}^{ST} = -1.194 + 1.081\hat{b}_{VE}^{ST}$$

*3.3.2. Duration of Tasks: Long vs. Short*

In factors of this experiment, task duration consist of long duration and short duration. For right forearm, the BPD result of long duration tasks, measured in RE, was denoted as $B_{RE}^{L}$, and that measured in VE was denoted as $B_{VE}^{L}$. The BPD result of short duration tasks, measured in RE, was denoted as $B_{RE}^{S}$, and that measured in VE was denoted as $B_{VE}^{S}$. Observed data of $B_{RE}^{L}$, $B_{VE}^{L}$, $B_{RE}^{S}$, and $B_{VE}^{S}$ were filtered by two eliminating rules mentioned above, and 27 data of each matrix were remained for analysis. The descriptive information of $B_{RE}^{L}$, $B_{VE}^{L}$, $B_{RE}^{S}$, and $B_{VE}^{S}$ was demonstrated in Table 10.

Table 10 The descriptive information of $B_{RE}^{L}$, $B_{VE}^{L}$, $B_{RE}^{S}$, and $B_{VE}^{S}$

| Duration of Tasks | BPD | Mean | Std Deviation |
|---|---|---|---|
| Long | $B_{RE}^{L}$ | 2.657 | 1.420 |
| | $B_{VE}^{L}$ | 3.745 | 1.251 |
| Short | $B_{RE}^{S}$ | 1.798 | 1.264 |
| | $B_{VE}^{S}$ | 2.889 | 1.012 |

Difference between $B_{RE}^{L}$ and $B_{VE}^{L}$ was tested with paired t-test. Their difference was significant ($t_{RE-VE} = -5.071$, $p < 0.05$). Furthermore $B_{RE}^{L}$ and $B_{VE}^{L}$ were regressed with linear model, and the result was ($R^2 = 0.433$, $F = 19.108$, $p < 0.05$):

$$\hat{b}_{RE}^{L} = -1.39 + 0.747\ \hat{b}_{VE}^{L}$$

In addition, difference between $B_{RE}^{S}$ and $B_{VE}^{S}$ was tested with paired t-test and its result was significant ($t_{RE-VE} = -5.990$, $p < 0.05$). Furthermore $B_{RE}^{S}$ and $B_{VE}^{S}$ were regressed with linear model and the result was ($R^2 = 0.473$, $F = 21.499$, $p < 0.05$):

$$\hat{b}_{RE}^{S} = -0.683 + 0.859\ \hat{b}_{VE}^{S}$$

*3.3.3. Size of target points: Large vs. Small*

In factors of this experiment, size of target points consists of large point and small point. For right forearm, the BPD result of large point tasks, measured in RE, was denoted as $B_{RE}^{LG}$, and that measured in VE was denoted as $B_{VE}^{LG}$. The BPD result of small point tasks, measured in RE, was denoted as $B_{RE}^{SL}$, and that measured in VE was denoted as $B_{VE}^{SL}$. Observed data of $B_{RE}^{LG}$, $B_{VE}^{LG}$, $B_{RE}^{SL}$, and $B_{VE}^{SL}$ were filtered by two eliminating rules mentioned above, and 27 data of each matrix were remained for analysis. The descriptive information of $B_{RE}^{LG}$, $B_{VE}^{LG}$, $B_{RE}^{SL}$, and $B_{VE}^{SL}$ was demonstrated in Table 11.

Table 11 The descriptive information of $B_{RE}^{LG}$, $B_{VE}^{LG}$, $B_{RE}^{SL}$, and $B_{VE}^{SL}$

| Size of target points | BPD | Mean | Std Deviation |
|---|---|---|---|
| Large | $B_{RE}^{LG}$ | 2.066 | 1.238 |
|  | $B_{VE}^{LG}$ | 3.241 | 1.032 |
| Small | $B_{RE}^{SL}$ | 2.416 | 1.465 |
|  | $B_{VE}^{SL}$ | 3.333 | 1.218 |

Paired t-test was performed for difference between $B_{RE}^{LG}$ and $B_{VE}^{LG}$, and its result was significant ($t_{RE-VE} = -6.103$, $p < 0.05$). Furthermore $B_{RE}^{LG}$ and $B_{VE}^{LG}$ were regressed with linear model and the result was ($R^2 = 0.391$, $F = 16.066$, $p < 0.05$):

$$\hat{b}_{RE}^{LG} = -0.364 + 0.750\, \hat{b}_{VE}^{LG}$$

In addition, difference between $B_{RE}^{SL}$ and $B_{VE}^{SL}$ was tested with paired t-test. The result was significant ($t_{RE-VE} = -4.937$, $p < 0.05$). Furthermore $B_{RE}^{SL}$ and $B_{VE}^{SL}$ were regressed with linear model. The result was ($R^2 = 0.571$, $F = 33.304$, $p < 0.05$):

$$\hat{b}_{RE}^{SL} = -0.614 + 0.909\, \hat{b}_{VE}^{SL}$$

*3.4. Rated perceiving exertion(RPE)*

The result of RPE measured in RE was denoted as $R'_{RE}$, and that measured in VE was denoted as $R'_{VE}$. The mean of $R'_{RE}$ was 11.729 (s.d.=1.778), and the mean of $R'_{VE}$ was 14.375 (s.d.=1.580). The difference between $R'_{RE}$ and $R'_{VE}$ was tested and the result was different significantly ($t_{RE-VE} = -6.345$, $p < 0.05$).

Significant difference of RPE between RE and VE were also found in different experimental configurations based on three factors: working posture tasks, duration of tasks, and size of target points (Table 12).

Table 12 Descriptive information of RPE

| Factor | Levels | RE | VE | $t_{RE-VE}$ |
|---|---|---|---|---|
| Working posture task | SD | 11.683(1.925) | 14.575(1.674) | -6.490** |
|  | ST | 11.775(1.890) | 14.283(1.748) | -6.088** |
| Duration of task | L | 12.375(2.037) | 15.075(1.652) | -7.140** |
|  | S | 11.083(1.780) | 13.783(1.606) | -5.648** |
| Size of target points | LG | 11.633(1.912) | 14.258(1.568) | -6.058** |
|  | SL | 11.818(1.848) | 14.600(1.636) | -6.728** |

Note: () is standard deviation, ** represents $p < 0.05$

In addition, experimental factor of duration tasks had effect on RPE. The mean of RPE with long duration tasks, denoted as $R^L$, was 13.680 (s.d.=1.599), and the mean of RPE with short duration tasks, denoted as $R^S$ was 12.409 (s.d.=1.068). Their difference was significant ($t_{L-S} = 5.871$, $p < 0.05$). However, for the factor of working posture task, difference of RPE between $R^{ST}$ and $R^{SD}$ was not significant ($t_{ST-SD} = -0.268$, $p = 0.791$). For the factor of size of target points, difference of RPE between $R^{LG}$ and $R^{SL}$ was not significant ($t_{LG-SL} = -1.622$, $p = 0.116$).

## 4. Discussion

In our experiment, different aspects of the same manual handling operations were evaluated using different evaluation methods both in RE and VE sessions. Posture and physical fatigue were used to evaluate the physical tasks objectively, while RPE and BPD were used to assess the tasks subjectively. Comparisons between these evaluations were made to check the availability of those evaluation methods in VE.

Slight differences were found in postures under different working environment, since 80.4% of the 240 observed differences were less than 5%. In this experiment, there were no restrict to limiting posture of subjects while performing tasks. In conventional ergonomics posture assessment methods, the flexible range of joints was divided into several segments. For examples, the total flexible range (360°) of elbow is divided into eight intervals in posture targeting method (Corlett, Madeley et al., 1979) . RULA and REBA divide the movement range of the forearm into two intervals: less than 60° and more than 100°, 60° - 100 ° (McAtamney and Nigel Corlett, 1993; Hignett and McAtamney, 2000). According to conventional posture assessment methods, postures of RE and VE can be treated as postures with same effect to people. While integrating posture-based evaluation methods in VE, the slight difference between VE and RE might not generate great differences while evaluating the postures. We think that it is feasible to use conventional observation methods in VE to evaluate physical tasks and almost the same evaluation results can be obtained in VE as in RE.

Significant differences were found in physical performances, BPD, and RPE evaluation results under difference environments. As shown in the all regression results of BPD, results in VE were greater than that in RE. These results indicated that subjects felt more discomfort in forearm in VE than in RE while performing the same tasks. In addition, the regression results of performance also indicated that subjects felt more fatigue in VE than RE. Moreover, RPE results give supports to the results of BPD and performance. RPE results indicated that more effort was required in VE than RE for same tasks. In addition, experimental factor of duration tasks had effect on RPE. The mean of RPE with long duration tasks (four loops) was higher than that with short duration tasks, and their difference was significant. This result indicates that long duration tasks require more effort than short duration tasks given by subjects.

In this experiment, subjects reported their sensation of presence in VE by presence questionnaire. The Cronbach $\alpha$ of the presence questionnaire was 0.815, and the mean of presence was 4.01 (s.d.=0.62) in the experiment. The result indicated that the VE could only provide an acceptable level of presence, but not immersive enough to replace the RE. Under ideal conditions, VR should provide 100% fidelity to subjects with immersive scenarios in VE as RE. However it is difficult to achieve 100% fidelity because of technical limitations, and these differences might generate influences on human's performance in VE.

ACT-R theory is used to analyze difference of evaluation results between RE and VE. An assembling task often consists of positioning, reaching, adjusting accurately, and hitting operations and it requires perception and manual action. An assembling task is processed by visual module and manual module of the mind according to ACT-R theory. The cognitive model of ACT-R theory divides generally the cognitive processing unit into three main parts: a visual module, a manual module, and a processing unit (Anderson, Bothell et al., 2004). The visual module estimates the position of the object and sends related information to processing unit of the mind. The processing unit selects an appropriate way and sends it to the manual module. Processing unit can select the most optimum way in RE because of practical experiences. It spends less time and generates less workload of people to accomplish operations. However, it is different when people recognize and estimate the position of a virtual object in VE. Because the perspective of visual scenes and position of viewpoint is different from that in RE, participant cannot determine the distance of virtual objects in VE accurately based on their experience of RE. Some researches indicated that subjects cannot determine the distance to them of an object accurately in VE with their experience of the RE (Arthur, Hancock et al., 1997; Witmer and Kline, 1998; Keyson, 2000; Armbrüster, Wolter et al., 2008). Then the procedure of coming from visual module to processing unit, then to manual module is required to repeat many times while subjects are performing their tasks in VE. The misestimation of the distance results in that subjects have to try and trail during their reaching, accurate adjusting, and hitting operations. Therefore, performing the same tasks requires additional effort in VE because of cognitive limitation in VE, and consequently causes more additional discomfort and fatigue. Furthermore, as observed in the experiment, subjects spent more time to finish a task in VE than RE.

Although significant differences were found in evaluation methods, linear regression method was used to assess the relationship between RE and VE as well, since it is believed that great linear correlation might be found when 100% fidelity could be provided in VE. After linear regression, $R^2$ in BPD and physical performance were 0.347 and 0.571, respectively. Based on the current technical specifications of the VR system, only with limited visual feedback, the regressed result indicates that evaluation in VE and RE was fairly related. Although the found linear regression results were not good enough, we could also state that certain relationship could be found if further technical improvements can be done in our future research work.

There are several technical limitations in this experiment. First, the presence provided by the VR simulation system is not good enough, and only acceptable presence was provided by the VR system. In addition, other feedbacks except visual feedback have not been involved in VE. Second, only a specific task was tested in our current research, and some other typical tasks have not been tested in our experiment. Third, subjects were not trained to how to use the VR devices for long time in this experiment, and their not enough acquaintance of VR devices might have effect on the accuracy of results.

## 5. Conclusion and Perspectives

In this experiment, the relationship and difference of subjective and objective evaluation results between RE and VE were analyzed for same tasks. Slight difference in postures engaged in manual handling operations was found, but those differences might not influence the evaluation results using posture-based evaluation methods. Significant differences were found both in objective (physical performance) and subjective evaluation methods (RPE and BPD). Results of BPD and performance in VE are greater than in RE for performing the same tasks partially due to extra cognitive efforts in VE. Furthermore, subjects felt more discomfort and fatigue in VE than RE. In spite of these differences, relationships of BPD and physical performance between VE and RE are analyzed with linear regression method. Correlations of BPD between RE and VE were found, for BPD and physical performance respectively, indicating that certain relationship might exist in evaluation results between RE and VE.

In our future work, multimodal feedbacks (e.g. audio, tactile) will be involved in the experiment to increase the presence in VE. Different evaluation methods will be carried out for the same tasks in VE in order to find out the trend of the correlation in evaluation results between VE and RE.

**Acknowledge**

The authors would like to acknowledge to the financial support form the National Natural Science Foundation under grand number 5020514, and from European Aeronautic Defense and Space (EADS). The project was carried out through collaboration between Tsinghua University and ECN promoted by Région des Pays de la Loire.